\documentclass[10pt, a4paper]{article}
\usepackage{lrec}
\usepackage{multibib}
\newcites{languageresource}{Language Resources}
\usepackage{graphicx}
\usepackage{tabularx}
\usepackage{soul}
\usepackage{stfloats}
\usepackage[dvipsnames]{xcolor}
\usepackage{booktabs}

\usepackage{hyperref}
\usepackage{times}
\usepackage{latexsym}

\usepackage{tikz}
\usepackage{amsmath,graphicx}
\usepackage{CJKutf8}
\usepackage[T1]{fontenc}
\usepackage{multirow}
\usepackage{amssymb}
\usepackage{subfig, graphicx}
\usepackage{url}
\usepackage{dialogue}

\usepackage[utf8]{inputenc}

\usepackage{capt-of}
\usepackage{varwidth}

\title{ASCEND: A Spontaneous Chinese-English Dataset for Code-switching in Multi-turn Conversation}

\name{Holy Lovenia$^\star$\thanks{$^\star$ These authors contributed equally.}, Samuel Cahyawijaya$^\star$, Genta Indra Winata$^\star$$^\dagger$\thanks{$^\dagger$ The work was done when the author was studying in The Hong Kong University of Science and Technology.}, Peng Xu, \\ \large{\textbf{Xu Yan, Zihan Liu, Rita Frieske, Tiezheng Yu, Wenliang Dai, Elham J. Barezi,}} \\ \large{\textbf{Qifeng Chen, Xiaojuan Ma, Bertram E. Shi, Pascale Fung}}}
\address{The Hong Kong University of Science and Technology \\
\texttt{\{hlovenia, scahyawijaya, giwinata\}@connect.ust.hk}
}

\abstract{
Code-switching is a speech phenomenon occurring when a speaker switches language during a conversation. Despite the spontaneous nature of code-switching in conversational spoken language, most existing works collect code-switching data from read speech instead of spontaneous speech. ASCEND (\textbf{A} \textbf{S}pontaneous \textbf{C}hinese-\textbf{En}glish \textbf{D}ataset) is a high-quality Mandarin Chinese-English code-switching corpus built on spontaneous multi-turn conversational dialogue sources collected in Hong Kong. We report ASCEND's design and procedure for collecting the speech data, including annotations. ASCEND consists of 10.62 hours of clean speech, collected from 23 bilingual speakers of Chinese and English. Furthermore, we conduct baseline experiments using pre-trained wav2vec 2.0 models, achieving a best performance of 22.69\% character error rate and 27.05\% mixed error rate.\\ \newline 
}

\begin{document}

\maketitleabstract

\section{Introduction}
\label{sec:intro}

% Overview Spontaneous vs Read Corpus
% Most of our knowledge about spoken language and speech processing comes from monolingual individuals producing scripted speech in a controlled settings. Monolingual speech allows for researchers to exercise tight control over the linguistic backgrounds of the speakers and the linguistic material (e.g. reading or repeating sounds, words, or sentences). While highly informative, these controlled monolingual speech samples are not representative of the spoken language in the real world. Multilingualism is the norm, not the exception, and individuals regularly make creative linguistic choices.

Most of our knowledge about speech recognition and speech generation technologies comes from 
% individuals producing 
monolingual read speech data collected in a controlled setting~\cite{panayotov2015librispeech}. Monolingual read speech data allows researchers to exercise tight control over the linguistic backgrounds of speakers and the linguistic material (e.g. reading or repeating sounds, words, or sentences). While being highly informative, these monolingual read speech samples do not capture particular actualities of spoken speech~\cite{howell1991prosodic,blaauw1994prosodic,batliner1995prosodic,li2002speechprosody,yang2013understanding,haynes2015spontaneous}.

Code-switching is a phenomenon typical of spoken speech, characterized by alternating use of more than one language. It may occur within a single utterance, which is known as intra-sentential code-switching, or between utterance boundaries, which is commonly referred to as inter-sentential code-switching. To address this phenomenon, code-switching corpora in many different languages pairs have been introduced, including but not limited to Indonesian-English~\cite{rizal-stymne-2020-evaluating}, Filipino-Spanish~\cite{bautista2004ph-en}, Latin-Irish~\cite{horst2017ir-la}, Spanish-English~\cite{garca2018sp-en}, Hindi-English~\cite{si2011hi-en,dey2014hi-en}, and Chinese-English \cite{lyu2010seame}.

% Characteristics of Spontaneous vs Read Corpus
Since code-switching occurs mostly during spontaneous conversational speech, building models using spontaneous speech should be more beneficial than using read speech.
%Aside from code-switching, there are many other differences between spontaneous speech and read speech utterances.
While the frequency of code-switching itself in read speech could be manually adjusted by modifying the transcription, spontaneous and read speech still have many other differences characterized by certain factors.
For instance, reduced spectral space and increased spectral variance are observed in Japanese spontaneous speech~\cite{nakamura2008differences}. Increased spectral variance has also been observed in French spontaneous speech~\cite{rouas2010comparison}. Other studies observe a reduction in phoneme duration~\cite{liu2010dialect} and word duration ~\cite{spilkov2010english} in spontaneous speech. Different patterns from the variance of GMM supervector are also shown to be able to discriminate spontaneous and read speech data~\cite{asami2014read}. Read speech possesses different acoustic properties, and reliance on them in code-switching task might lead to a distributional shift, which consequently compromises the overall performance of the acoustic model in a real-setting. For this reason, building code-switching ASR using spontaneous speech is preferable to read speech.
%All these differences in spontaneous and read speech might lead to a distributional shift which will impact the overall performance of the acoustic model in the real-setting of a code-switching speech utterance.

In this work, we introduce ASCEND\footnote{We release ASCEND at \url{https://huggingface.co/datasets/CAiRE/ASCEND}.}, a spontaneous multi-turn conversational dialogue Mandarin Chinese-English code-switching corpus, to bridge the gap between the real-setting of code-switching speech utterances and the existing code-switching speech corpora. ASCEND comprises 10.62 hours of clean spontaneous Chinese-English code-switching data collected from dialogues between two people. To allow more variety in the utterances, speakers are diversified based on their English proficiency level and their Chinese dialects, covering Hong Kong, Taiwan, and various regions in Mainland China. In order to build a rich and diverse vocabulary, dialogues on various topics are incorporated into the corpus. These include education, persona, philosophy, sports, and technology. Overall, we collect 26 dialogue sessions with a total of 23 speakers. Our corpus is equally split between the genders.

\section{Related Work}
\label{sec:related-work}
% Spontaneous Speech Corpus and its effect
Code-switching has been widely studied in both text and speech modalities for multiple language pairs: 1) code-switching in Hindi-English, Bengali-English, Gujarati-English, and Tamil-English ~\cite{banerjee2018dataset}; 2) code-switching in Spanish-English and Modern Standard Arabic-Egyptian~\cite{calcs2018shtask}, Irish-Latin code-switching~\cite{lynn2019irish}; 3) code-switching in Arabic-English and Arabic-French~\cite{chowdhury2021omra}; and 4) code-switching in Chinese-English~\cite{lin2021bitod,lyu2010seame}. Furthermore, many solutions specific to code-switching have been established, such as multitask and meta learning for
code-switching~\cite{yu2020preliminary,song2017multitask,winata2018multitaskcs}, code-switched data augmentation method~\cite{qin2020cosdamlmc,winata2019code}, and adaptation method from large multilingual models for code-switching setting~\cite{winata2021multics,winata2021multilingual}.

Despite the gradual progression of code-switching research, existing code-switching solutions merely reach a decent level of performance, which is several times inferior to that of their monolingual counterparts, especially in the automatic speech recognition (ASR) task. For instance, for the traditional ASR task, word error rate (WER) of $\sim$2\%~\cite{anmol2020conformer,baevski2020wav2vec} and character error rate (CER) of $\sim$5\%~\cite{zhang2020unifiedsa} have been reported for Librispeech (English)~\cite{panayotov2015librispeech} and AiShell-1 (Chinese)~\cite{bu2017aishell} corpora, respectively. Code-switching ASR, on the other hand, has much poorer state-of-the-art performance, with 24.2\% mixed error rate (MER)~\cite{li2019integrating} and 29.30\% CER~\cite{winata2020mtl} for Chinese-English, 26.4\% WER for Arabic-English~\cite{chowdhury2021omra}, and 37.70\% WER for Arabic-French~\cite{chowdhury2021omra}. We argue that these performance gaps occur due to the limitation of existing code-switching corpora in comparison with monolingual corpora, notably for high-resource languages, e.g., English and Mandarin Chinese.

\begin{table}[t]
    \centering
    \begin{tabular}{c p{0.7\linewidth}}
        \toprule
        \textbf{Topic} & \textbf{Sample question} \\
        \toprule
        \multirow{2}{*}{Technology} & \multicolumn{1}{p{0.7\linewidth}}{\begin{CJK*}{UTF8}{gbsn}你使用任何社交媒体吗？\end{CJK*}} \\
                                    & \multicolumn{1}{p{0.7\linewidth}}{(Do you use any social media?)} \\
        \midrule
        \multirow{2}{*}{Sports} & \multicolumn{1}{p{0.7\linewidth}}{\begin{CJK*}{UTF8}{gbsn}是谁鼓励你参加这项运动？\end{CJK*}} \\
                                & \multicolumn{1}{p{0.7\linewidth}}{(Who inspired you to play this sport?)} \\
        \midrule
        \multirow{2}{*}{Education} & \multicolumn{1}{p{0.7\linewidth}}{\begin{CJK*}{UTF8}{gbsn}你们的course project是什么？\end{CJK*}} \\
                                    & \multicolumn{1}{p{0.7\linewidth}}{(What is your course project?)} \\
        \midrule
        \multirow{3}{*}{Philosophy} & \multicolumn{1}{p{0.7\linewidth}}{\begin{CJK*}{UTF8}{gbsn}你听说过火车电车问题吗？\end{CJK*}} \\
                                & \multicolumn{1}{p{0.7\linewidth}}{(Have you heard of the train trolley problem?)} \\
        \bottomrule
    \end{tabular}
    \caption{Examples of topic ideas and questions for the conversation in Session 2--4.}
    \label{tab:topic_examples}
\end{table}

% Spontaneous Speech Corpus and its effect
% As the demand of Chinese-English speech recognition has been rising,
In recent years, many speech corpora for Chinese-English code-switching have been introduced. CECOS corpus~\cite{shen2011cecos} is a collection of 12.1 hours of read Chinese-English code-switching corpus by Taiwanese speakers. SEAME corpus~\cite{lyu2010seame} consists of 30 hours of spontaneous intra-sentential code-switching speech utterances collected from 92 speakers, covering Chinese-English code-switching within Singaporean and Malaysian populations. OC16-CE80~\cite{wang2016mixlingual} is a Chinese-English code-switching corpus that consists of 80 hours of read speech collected from more than 1,400 speakers from the Mainland Chinese population. ASRU 2019~\cite{shi2020asru} is a large-scale Chinese-English code-switching corpus with 740 hours of utterances, 240 hours of Chinese-English code-switching  read utterances and 500 hours of monolingual Chinese utterances.\footnote{The paper makes no explicit mention whether the code-switching corpus uses read speech. We gathered this information from the competition website \url{https://www.datatang.com/competition}. There is also no indication of whether the Chinese corpus is read or spontaneous. (Access date: 12 November 2021)} The dataset is collected from multiple speakers from 30 provinces in Mainland China. Finally, ~\cite{li2012mandarin} introduce 36 hours of spontaneous Chinese-English code-switching speech recordings, mainly in Chinese, English, and Cantonese with a small proportion of German and French. They report that only a fraction of this corpus is transcribed.

Despite the abundance of Chinese-English code-switching resources, many are no longer publicly available. OC16-CE80 and ASRU 2019 were only available for past competition purposes and are no longer publicly available.\footnote{Some steps of the procedures required to obtain the dataset given by the affiliated institution are missing.} Moreover, CECOS and~\cite{li2012mandarin} are also no longer publicly available.\footnote{Dataset status was confirmed by contacting the authors and/or the affiliated institution.} Hence, there is no publicly available Chinese-English code-switching corpus as of now, except for SEAME.

\begin{table}[t]
    \centering
    \begin{tabular}{ccc}
    \toprule
    \textbf{Session} & \textbf{Average duration (minutes)} & \textbf{Done by} \\
    \midrule
    1 & 11.07 & 13 pairs \\
    2 & 13.78 & 13 pairs \\
    3 & 14.45 & 13 pairs \\
    4 & 13.85 & 10 pairs \\
    \bottomrule
    \end{tabular}
    \caption{Statistics of each recording session of our ASCEND corpus.}
    \label{tab:session-details}
\end{table}

\section{Corpus Collection}
\label{sec:corpus-collection}

In this section, we describe the setup and procedure for the audio recording used for collecting ASCEND's multi-turn conversational code-switched speech dialogues.

% In this section we describe the overview of our ASCEND corpus, the setups of the audio recording, and the recording guideline that is used for collecting the multi-turn conversational code-switched speech dialogue of our ASCEND corpus.

\subsection{Recording setup}
\label{sec:recording-setup}

ASCEND is collected through recording an informal conversation between two speakers. The recordings are made in a quiet classroom. Both speakers are seated across one another at a distance of $\sim$1 meter. Each speaker is equipped with a RODE SmartLav+ clip microphone as the recording device. The microphone is mounted on the speaker's shirt collar. The audio recording is set to a mono channel with a sample rate of 16 kHz, and the audio signal is encoded as 16-bit pulse-code modulation (PCM), producing a total bit rate of 256 kbps. The resulting audio file is stored in an uncompressed WAVE (.wav) file format.

\subsection{Recording procedure}
\label{sec:recording-procedure}

We collect the conversational audio recording data in the form of a casual one-on-one conversation. Both speakers take turns to ask a question, answer, or talk about a certain topic however they would like to, maintaining the natural course of the conversation. Short pauses, coughs, laughter, incomplete sentences, and other spontaneous responses that usually do not come up in a formal or organized setting are allowed in the conversation. Both speakers are encouraged to use code-switching at all times during the recording, as long as the utterance feels natural to the speaker. The task description, along with written consent and information that the conversation will be recorded and the resulting audio data published, is provided to all speakers before the recording begins.

The recording is split into several sessions. During the first session, both speakers get to know each other by exchanging information on personal topics, such as nicknames, family, favorite pastimes, and recent activities. This session is intended to gradually make them feel at ease around one another to spark a more interactive and dynamic conversation in upcoming sessions. In the next two or three (depending on the remaining time) sessions, the conversation takes off on a broader subject to encourage a larger variety of vocabulary. To facilitate this, we provide a list of topic ideas and questions for the speakers to gather inspiration from. A few examples from this list can be seen in Table \ref{tab:topic_examples}. For each session, speakers can choose one topic they are comfortable with and begin the conversation based on it. To ensure a natural conversation flow, no restriction is enforced to keep the conversation in-topic; speakers are free to deviate from the determined topic at any point of the conversation.

\begin{table}[t]
    \centering
    \begin{tabular}{c|c}
    \toprule
    \textbf{\# Speakers} & 23 speakers \\
    \textbf{\# Sessions} & 49 sessions \\
    \textbf{\# Raw recordings} & 98 recordings \\
    \textbf{Avg. utterances} & 128.27 per speaker per session \\
    \textbf{English speaking rate} & 152.31 words/minute \\
    \textbf{Chinese speaking rate} & 262.33 characters/minute \\
    \bottomrule
    \end{tabular}
    \caption{The overview of the collected raw audio data of our ASCEND corpus.}
    \label{tab:overview-raw-data}
\end{table}

\begin{table*}[b!]
\begin{minipage}[ht]{0.3\linewidth}
    \centering
    \includegraphics[width=\linewidth]{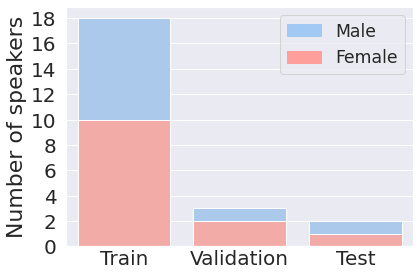}
    \captionof{figure}{Speaker split in ASCEND corpus.}
    \label{fig:speaker-split}
\end{minipage}
\hfill
\begin{minipage}[ht]{0.7\linewidth}
    \centering
    \begin{tabular}{lcccccccc}
        \toprule 
        \multirow{2}{*}{\textbf{Gender}} & \multicolumn{4}{c}{\textbf{\# Utterance}} & \multicolumn{4}{c}{\textbf{Duration (hr)}} \\
        \cmidrule(l{2pt}r{2pt}){2-5} \cmidrule(l{2pt}r{2pt}){6-9}
        & \multicolumn{1}{l}{\textbf{Train}} & \multicolumn{1}{l}{\textbf{Val}} & \multicolumn{1}{l}{\textbf{Test}} & \textbf{Total} & \multicolumn{1}{l}{\textbf{Train}} & \multicolumn{1}{l}{\textbf{Val}} & \multicolumn{1}{l}{\textbf{Test}} & \textbf{Total} \\ \midrule
        \textbf{Female} & 4,591 & 484 & 861 & 5,936 & 4.04 & 0.46 & 0.48 & 4.98 \\
        \textbf{Male} & 5,278 & 646 & 454 & 6,378 & 4.74 & 0.46 & 0.44 & 5.63 \\
        \midrule
        \textbf{Total} & 9,869 & 1,129 & 1,315 & 12,314 & 8.78 & 0.92 & 0.92 & 10.62 \\
        \bottomrule
    \end{tabular}
    \caption{Train, validation, and test split in ASCEND corpus.}
    \label{dataset-split}
\end{minipage}
\end{table*}

The recording takes approximately one hour. It includes 5 minutes of instructions, 40--55 minutes of mixed-language conversation and breaks in between each session. We record 13 casual one-on-one conversations with 13 speaker pairs, collecting a total of 49 sessions (Table \ref{tab:session-details}). Three speakers participate twice, with a different conversation partner in each round. On average, the first session goes on for 11 minutes, while the later sessions for around 14--15 minutes. For each session, we obtain two recordings from each speaker, which sum to 98 raw audio files. Table \ref{tab:overview-raw-data} presents the overall statistics of the raw data collected from our ASCEND corpus.

\section{Annotation}
\label{sec:annotation}

The raw audio recordings of the sessions are split into utterances by a professional annotation company based on a natural semantic boundary or a long pause between utterances. Utterances corresponding to a speaker are obtained from the audio file recorded by the respective microphone. The utterances are manually transcribed in Chinese characters, English letters, or a mix of both, depending on the language in use. For consistency and accuracy of the annotation results, we formulate guidelines for the transcription annotation, as follows:

\begin{enumerate}
    \item Numbers are written as words instead of numerals. For example, "24 hours" is transcribed as "twenty four hours" in the corpus. 
    \item Abbreviations are transcribed as capital letters or separated by a space.
    \item Contractions and shortened versions of words (e.g., "can't", "won't", and "it's") are not expanded. We keep contractions as-is because of the possible difference in phoneme.
    \item Fillers or discourse particles are annotated as either: \textit{ah}, \textit{oh}, or \textit{um}.
    \item Punctuation symbols, such as period (.), comma (,), question mark (?), and exclamation mark (!), are not used to mark the utterances.
    \item Unintelligible speech is marked with an \texttt{[UNK]} placeholder token.
    \item Repetitions are preserved. Annotators write the words down as what they hear from the speech data. For example, "I don't (I don't) think they should be in the Olympic Games" is transcribed as "I don't I don't think they should be in the Olympic Games".
\end{enumerate}

\paragraph{Post-annotation processing}

To ensure the quality of our speech data, we do a second round of processing with a mix of manual and automatic checking. We inspect the transcriptions and remove the unnecessary symbols, whitespace, and annotation inconsistency. We exclude utterances that only contain \texttt{[UNK]} from the corpus. We re-format utterance audio files that do not follow the recording standards mentioned in Section \ref{sec:recording-setup}

\paragraph{Corpus splitting}

\begin{figure}[t]
    \centering
    \includegraphics[width=1.0\linewidth]{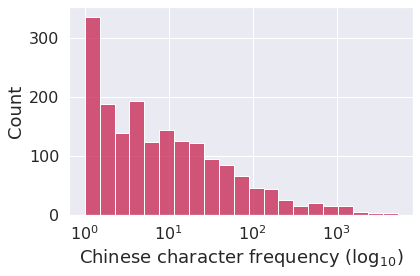}
    \caption{The distribution of the log character frequency in Chinese speech in ASCEND.}
    \label{fig:zh-log-freq}
\end{figure}

We divide the utterances into train, validation, and test sets. The sets have disjoint combinations of speakers (as presented by Figure \ref{fig:speaker-split}) to enable this corpus to be used for the speaker-independent speech recognition task. Within each split, we balance the total duration of the audio data from each gender. At the end of this process, ASCEND is formed with the approximate ratio of 8:1:1 for its train, validation, and test sets respectively. This ratio is derived from both the audio duration and the number of utterances in each split. Table \ref{dataset-split} describes the statistics of ASCEND's train, validation, and test sets.

\section{ASCEND: A Spontanenous Chinese-English Dataset}
\label{sec:ascend-corpus}

In this section, we report statistical findings regarding the Chinese-English code-switching of ASCEND. We also provide the statistics of the speakers who have participated in the corpus collection.

\subsection{Corpus profile}
\label{sec:corpus-profile}

\begin{table}[b]
    \centering
    \begin{tabular}{rcc}
        \toprule
        \textbf{Language} & \textbf{\# Utterance} & \textbf{Duration (hr)} \\
        \midrule
        \textbf{Chinese} (49.85\%) & 6,139 & 5.32 \\
        \textbf{English} (23.14\%) & 2,850 & 2.42 \\
        \textbf{Mixed} (27.01\%) & 3,325 & 2.88 \\
        \bottomrule
    \end{tabular}
    \caption{Utterance distribution per language.}
    \label{tab:lang-usage-in-dataset}
\end{table}

\begin{figure}[t]
    \centering
    \includegraphics[width=1.0\linewidth]{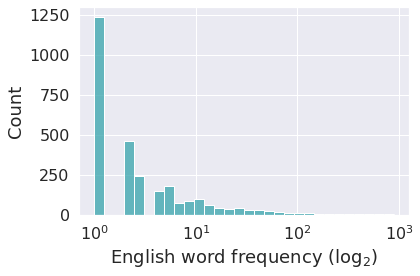}
    \caption{The distribution of the log word frequency in English speech in ASCEND.}
    \label{fig:en-log-freq}
\end{figure}

ASCEND comprises 10.62 hours and $\sim$12.3K utterances of spontaneous speech, with an average duration of 3.10 seconds per utterance. ASCEND includes a total of 145,146 tokens (i.e., words in English and characters in Chinese) with 1,795 types of Chinese characters and 2,860 types of English words. An utterance is approximately 11.78 tokens long. In both languages, we find that a small portion of the vocabulary (e.g., particles, pronouns, affirmations, etc.) appears much more frequently than the rest. The distribution of the token frequency in ASCEND is depicted in Figure \ref{fig:zh-log-freq} and Figure \ref{fig:en-log-freq}.

ASCEND is collected from multiple speakers from different locations, including Taiwan, Hong Kong, and various provinces in Mainland China. Section \ref{sec:speaker-distribution} will discuss more details about our speakers. In terms of code-switching characteristics, the dialogues in ASCEND encompass both inter-sentential code-switching (from monolingual Chinese to monolingual English utterance or vice versa) and intra-sentential code-switching (mixed Chinese-English). Table \ref{tab:lang-usage-in-dataset} describes the proportion of the languages used in the speech data.

\subsection{Speaker distribution}
\label{sec:speaker-distribution}

We hire 23 university students as our speakers, all of whom are native Chinese speakers who converse using English on a daily basis. Their personal information is obtained from an online form we provide during the speaker registration. Of the speakers, 13 identify as female and the other 10 identify as male. The speakers' ages range from 19 to 30 years old, with a mean of 24 and a standard deviation of 2.24.

\begin{table}[b]
    \centering
    \begin{tabular}{c c c c}
        \toprule
        \textbf{English study} & \textbf{Chinese} & \textbf{English} & \textbf{Mixed} \\
        \midrule
        < 10 years & 53.38\% & 23.78\% & 22.84\% \\
        10-15 years & 50.57\% & 21.07\% & 28.37\% \\
        > 15 years & 47.87\% & 27.77\% & 24.37\% \\
        \bottomrule
    \end{tabular}
    \caption{Language usage by years of English study.}
    \label{tab:lang-usage-by-en-study}
\end{table}

In addition to gender and age demographics, we also collect information that is indicative of their English proficiency to ensure the quality of the acquired code-switching utterances (Table \ref{tab:lang-usage-by-en-study}). Most speakers have been studying English for 10 years or more, except for two whose experience has just passed the five year count. We also collect their speaking scores (according to IELTS or TOEFL iBT) to measure their fluency in English as a second language. The speaking scores among the speakers are then standardized to IELTS band criteria. We find that all speakers reach or surpass the 5.5 mark, with an average score of 6.5.

\subsection{Topic and code-switching}
\label{sec:topic-and-code-switching}

\begin{table}[b]
    \centering
    \begin{tabular}{c c c c}
        \toprule
        \textbf{Topic} & \textbf{Chinese} & \textbf{English} & \textbf{Mixed} \\
        \midrule
        Education & 51.57\% & 23.16\% & 25.27\% \\
        Persona & 48.76\% & 25.85\% & 25.40\% \\
        Philosophy & 44.91\% & 26.54\% & 28.55\% \\
        Sports & 55.22\% & 21.85\% & 22.94\% \\
        Technology & 48.06\% & 20.01\% & 31.93\% \\
        \bottomrule
    \end{tabular}
    \caption{Language usage by conversation topic.}
    \label{tab:lang-usage-by-conv-topic}
\end{table}

As mentioned in Section \ref{sec:recording-procedure}, each session uses one topic as a conversation starter. The topic in the first session is always persona, which covers both speakers' backgrounds such as name, hobbies, and age. The topics for the later sessions adhere to the speakers' choice, which is either education, philosophy, sports, or technology. From the total of 49 sessions, 12 correspond to education, 13 correspond to persona, 4 correspond to philosophy, 7 correspond to sports, and 13 correspond to technology.

In general, around half of the utterances (44.78\%--55.09\%) spoken for all of the topics consists of code-switching. Although the proportions of utterances with inter-sentential and the ones with intra-sentential code-switching are quite balanced, as shown in Table \ref{tab:lang-usage-by-conv-topic}, the usage of intra-sentential code-switches increases for topics involving many widely-known English terms. One of these topics is technology, where intra-sentential code-switching makes up 31.93\% of the utterances. The other is philosophy, which is composed of the highest overall percentage of code-switching. We also find that, despite code-switches using monolingual English utterances tending to be more occasional, their occurrence frequency increases along with the speakers' familiarity and knowledge about the conversation subject. For example, talking about communication devices during the technology topic or themself during the persona topic triggers inter-sentential code-switches slightly more often among the speakers.

\subsection{Common English phrases used in ASCEND}

\begin{table}[b]
    \centering
    \resizebox{1.0\linewidth}{!}{
        \begin{tabular}{clll}
            \toprule
            \multirow{2}{*}{\textbf{Top}} & \multicolumn{3}{p{0.7\linewidth}}{\textbf{English phrases}} \\
            & \multicolumn{1}{p{0.2\linewidth}}{\textbf{1-gram}} & \multicolumn{1}{p{0.2\linewidth}}{\textbf{2-gram}} & \multicolumn{1}{p{0.3\linewidth}}{\textbf{3-gram}} \\
            \midrule
            1 & the & do you & do you think \\
            2 & you & in the & what do you \\
            3 & to & you can & how to say \\
            4 & like & kind of & in hong kong \\
            5 & and & smart phone & this kind of \\
            6 & is & to do & you want to \\
            7 & in & hong kong have & so do you \\
            8 & so & you have & want to do \\
            9 & of & want to & you are not \\
            10 & for & you know & meaning of life \\
            \bottomrule
        \end{tabular}
    }
    \caption{Top 10 English 1-gram, 2-gram, and 3-gram phrases.}
    \label{tab:top-en-phrases}
\end{table}

\begin{figure*}[t]
\begin{minipage}[b]{0.3\linewidth}
    \centering
    \includegraphics[width=\linewidth]{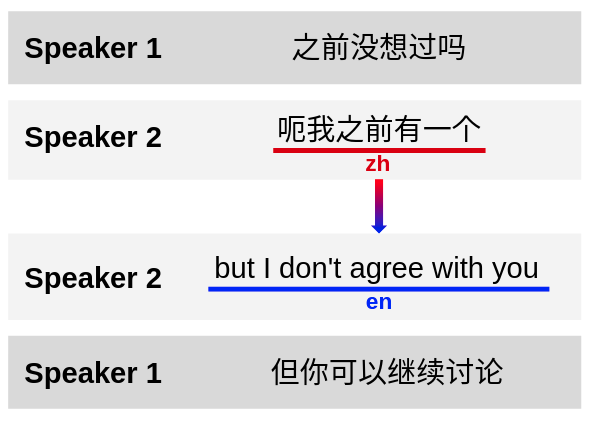}
    \captionof{figure}{Example of inter-sentential code-switching in ASCEND.}
    \label{fig:inter-switch}
\end{minipage}
\hfill
\begin{minipage}[b]{0.33\linewidth}
    \centering
    \includegraphics[width=\linewidth]{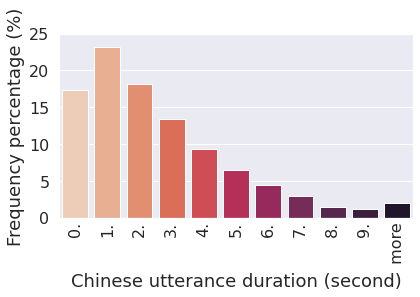}
    \captionof{figure}{Distribution of Chinese utterance duration in ASCEND.}
    \label{fig:zh-utt-dur-dist}
\end{minipage}
\hfill
\begin{minipage}[b]{0.33\linewidth}
    \centering
    \includegraphics[width=\linewidth]{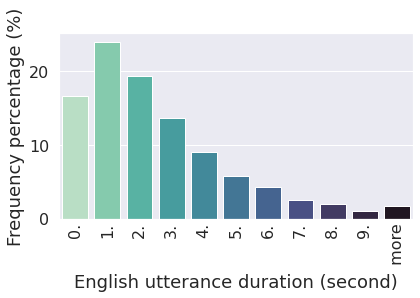}
    \captionof{figure}{Distribution of English utterance duration in ASCEND.}
    \label{fig:en-utt-dur-dist}
\end{minipage}
\end{figure*}

While the lexical resources used during code-switching from Chinese to English vary, some come up more frequently than others in the conversations. According to Table \ref{tab:top-en-phrases}, the types of phrases that often occur in our corpus are related to asking a question (e.g., "do you think" and "what do you") and giving or thinking of a response (e.g., "how to say", "want to do", and "you know"). Aside from these, speakers quite frequently exchange phrases that are used to describe an idea (e.g., "like", "in the", "you can", and "this kind of"). A few topic-related phrases, such as "smart phone" for technology and "meaning of life" for philosophy, are also mentioned a lot during the discussions.

\subsection{Inter-sentential code-switching in ASCEND}

Our ASCEND corpus contains several inter-sentential code-switching instances. Inter-sentential code-switching differs from intra-sentential in a way that its language switch occurs between utterances. For example, in Figure \ref{fig:inter-switch}, the second speaker completes the first utterance in Chinese then switches to English for the entire second utterance. As a result, all the involved utterances are still monolingual despite a language switch occurring. As shown by Figure \ref{fig:zh-utt-dur-dist} and Figure \ref{fig:en-utt-dur-dist}, we find that the monolingual utterances in ASCEND have a similar duration distribution for both Chinese and English utterances.

\subsection{Intra-sentential code-switching in ASCEND}

\begin{figure}[h]
    \centering
    \includegraphics[width=\linewidth]{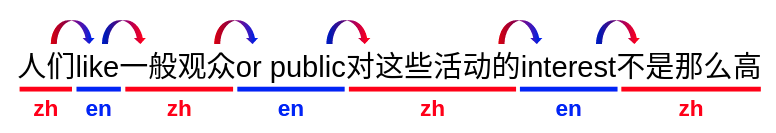}
    \caption{Intra-sentential code-switching utterance with six language turns.}
    \label{fig:utt-6-lang-turns}
\end{figure}

Aside from inter-sentential code-switching, ASCEND also consists of numerous intra-sentential code-switching utterances. An utterance is considered to have intra-sentential code-switching when a switch from one language to another happens within the utterance at least once. We refer to this language switching phenomenon as language turn. In the example in Figure \ref{fig:utt-6-lang-turns}, the utterance begins in Chinese, switches to English, goes back to Chinese, and so on until the language turns sum to six. In practice, most utterances tend to have a lower number of language turns. In the intra-sentential code-switching utterances in our ASCEND corpus, language turns appear 2.18 times per utterance on average, with a maximum of 14 times in a single utterance.

\begin{table}[t]
    \centering
    \begin{tabular}{cll}
        \toprule
        \multirow{2}{*}{\textbf{Top}} & \multicolumn{2}{p{0.7\linewidth}}{\textbf{Language turn}} \\
                                      & \multicolumn{1}{p{0.35\linewidth}}{\textbf{zh $\rightarrow$ en}} & \multicolumn{1}{p{0.35\linewidth}}{\textbf{en $\rightarrow$ zh}} \\
        \midrule
        1 & \begin{CJK*}{UTF8}{gbsn}个\end{CJK*} project & school \begin{CJK*}{UTF8}{gbsn}的\end{CJK*} \\
        2 & \begin{CJK*}{UTF8}{gbsn}读\end{CJK*} phd & phd \begin{CJK*}{UTF8}{gbsn}的\end{CJK*} \\
        3 & \begin{CJK*}{UTF8}{gbsn}个\end{CJK*} topic & ok \begin{CJK*}{UTF8}{gbsn}的\end{CJK*} \\
        4 & \begin{CJK*}{UTF8}{gbsn}做\end{CJK*} research & smartphone \begin{CJK*}{UTF8}{gbsn}的\end{CJK*} \\
        5 & \begin{CJK*}{UTF8}{gbsn}的\end{CJK*} major & phone \begin{CJK*}{UTF8}{gbsn}的\end{CJK*} \\
        \bottomrule
    \end{tabular}
    \caption{Top 5 code-switches in language turns between Chinese and English.}
    \label{tab:top-lang-turns}
\end{table}

\begin{figure*}[b]
\begin{minipage}[b]{0.45\linewidth}
    \centering
    \includegraphics[width=\linewidth]{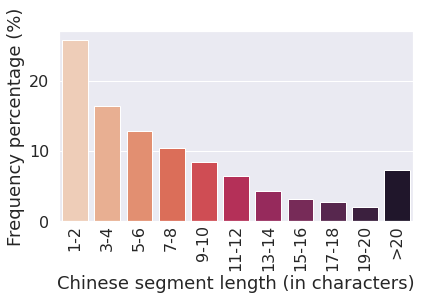}
    \captionof{figure}{The distribution of Chinese segment length.}
    \label{fig:zh-segment-dist}
\end{minipage}
\hfill
\begin{minipage}[b]{0.45\linewidth}
    \centering
    \includegraphics[width=\linewidth]{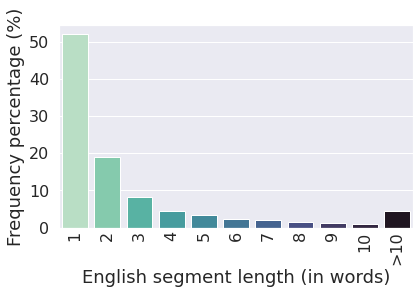}
    \captionof{figure}{The distribution of English segment length.}
    \label{fig:en-segment-dist}
\end{minipage}
\end{figure*}

\paragraph{Language turn within utterances}

As the speech data in our ASCEND corpus is spontaneous, all code-switches, including those used in language turns, occur on the speaker's own accord without any fixed or predefined rules. Nevertheless, people tend to follow certain lexical patterns during code-switching, so a few mixes of Chinese and English phrases get used in language turns more frequently than others. We select one Chinese character and one English word from every language turn and sort them based on their occurrence frequency. Table \ref{tab:top-lang-turns} reports the five most common language turns for code-switching from Chinese to English and vice versa in ASCEND.

\paragraph{Utterance as multiple monolingual segments}

As shown in Figure \ref{fig:utt-6-lang-turns}, the presence of language turns causes the corresponding intra-sentential code-switching utterance to be composed of multiple monolingual segments. Depending on the language usage and the speaker, these segments vary in length. To observe the style of intra-sentential code-switching in spontaneous conversations, we separate the Chinese segments from the English ones, then calculate the number of segments found in each utterance. We find that an intra-sentential code-switching utterance typically comprises 1.75 Chinese segments and 1.38 English segments. In addition to the number of segments, we also calculate the occurrence frequency for each segment length. We report the overall distribution of the number of Chinese characters per segment in Figure \ref{fig:zh-segment-dist} and the number of English words per segment in Figure \ref{fig:en-segment-dist}.

\begin{table}[t]
    \centering
    \begin{tabular}{cllll}
        \toprule
        \multirow{2}{*}{\textbf{Top}} & \multicolumn{2}{p{0.325\linewidth}}{\textbf{Chinese segments}} & \multicolumn{2}{p{0.35\linewidth}}{\textbf{English segments}} \\
                                      & \multicolumn{1}{p{0.15\linewidth}}{\textbf{1-char}} & \multicolumn{1}{p{0.175\linewidth}}{\textbf{2-char}} & \multicolumn{1}{p{0.15\linewidth}}{\textbf{1-word}} & \multicolumn{1}{p{0.2\linewidth}}{\textbf{2-word}} \\
        \midrule
        1 & \begin{CJK*}{UTF8}{gbsn}的\end{CJK*} & \begin{CJK*}{UTF8}{gbsn}就是\end{CJK*} & ai & smart phone \\
        2 & \begin{CJK*}{UTF8}{gbsn}啊\end{CJK*} & \begin{CJK*}{UTF8}{gbsn}然后\end{CJK*} & phd & social media \\
        3 & \begin{CJK*}{UTF8}{gbsn}是\end{CJK*} & \begin{CJK*}{UTF8}{gbsn}所以\end{CJK*} & ok & hong kong \\
        4 & \begin{CJK*}{UTF8}{gbsn}对\end{CJK*} & \begin{CJK*}{UTF8}{gbsn}这个\end{CJK*} & so & i think \\
        5 & \begin{CJK*}{UTF8}{gbsn}吗\end{CJK*} & \begin{CJK*}{UTF8}{gbsn}那个\end{CJK*} & and & it's like \\
        \bottomrule
    \end{tabular}
    \caption{Top 5 short monolingual segments in intra-sentential code-switching.}
    \label{tab:top-short-monolingual-segments}
\end{table}

Despite having a similar number of segments in an utterance, the characteristics of the Chinese segment length distribution differs from the English, with the former having a more even length distribution than the latter. Short Chinese segments (1--4 characters long) make up approximately 35\% of the population, while the percentage doubles for English. Around 70\% of English segments found in intra-sentential code-switching utterances consist of one or two words. Although language turns can occur in both languages, we can see that people tend to speak in longer Chinese segments (7.96 characters per segment on average) then switch to a shorter English segment (2.96 words per segment on average) in between. This phenomenon is expected, considering that all the speakers have Chinese as their first language. This speaking pattern aligns with the characteristics of code-switching in Hong Kong, Taiwan, Singapore, and Malaysia reported by 
\cite{chan2005development}, \cite{lyu2006speech}, and \cite{lyu2010seame}. English-dominated utterances with Chinese code-switches also appear in ASCEND, albeit more occasionally. Table \ref{tab:top-short-monolingual-segments} presents one-token and two-token segments that are commonly utilized as code-switches in ASCEND.

\section{Baseline Experiment}

In this section, we conduct an experiment on ASCEND to show its reliability and validity as a code-switching speech corpus.\footnote{The baseline experiment code can be found at \url{https://github.com/HLTCHKUST/ASCEND}.} For the experiment, a state-of-the-art speech recognition model architecture, namely wav2vec 2.0~\cite{baevski2020wav2vec}, is employed. As no code-switching ASR model is available, we utilize wav2vec 2.0 models with different initializations as the baselines. The first two models are pre-trained on the English corpus and the Chinese corpus of Common Voice, respectively. Motivated by~\cite{winata2021multics} who use multilingual models to approach the code-switching task, the third model is initialized with a multilingual wav2vec 2.0 pre-trained on 53 languages of the Common Voice corpus.

% For the experiment, we employ a state-of-the-art model architecture, wav2vec 2.0 ~\cite{}. Specifically, we employ a fine-tuned version of wav2vec 2.0 model that is already adapted to the ASR task. As there is no code-switching ASR model available, we instead employ two versions of ASR wav2vec 2.0 models, one fine-tuned on the English CommonVoice dataset and the one on the Chinese CommonVoice dataset.

% Baseline models -- maybe no need for this section?

% Implementation details [WIP]

\paragraph{Preprocessing.} Before we fine-tune both models on ASCEND, we omit unnecessary characters and symbols from the transcription data. The resulting texts are used to build ASCEND-specific vocabulary, which we leverage to extend the pre-trained tokenizer that comes with the model. Table \ref{tab:vocab-size} shows the vocabulary size of each model with and without ASCEND-specific vocabulary.

\begin{table}[t]
    \centering
    \resizebox{1.0\linewidth}{!}{
        \begin{tabular}{cllll}
            \toprule
            \multirow{2}{2 cm}{\centering\textbf{Pre-training language}} & \multicolumn{2}{c}{\textbf{Vocabulary size}} \\
                & \multicolumn{1}{c}{\textbf{Pre-trained only}} & \multicolumn{1}{c}{\textbf{With ASCEND}} \\
            \midrule
            Chinese & 3503 & 3593 \textcolor{Green}{(+90)} \\
            English & 33 & 1833 \textcolor{Green}{(+1800)} \\
            Multilingual & 9913 & 9920 \textcolor{Green}{(+7)} \\
            \bottomrule
        \end{tabular}
    }
    \caption{Vocabulary size of the models before and after the additions from ASCEND}
    \label{tab:vocab-size}
\end{table}

\begin{figure*}[b!]
\begin{minipage}[b]{0.32\linewidth}
    \centering
    \includegraphics[width=\linewidth]{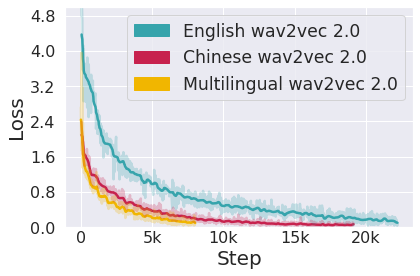}
    \captionof{figure}{Loss on ASCEND train set in the baseline experiments.}
    \label{fig:exp-train-loss}
\end{minipage}
\hfill
\begin{minipage}[b]{0.32\linewidth}
    \centering
    \includegraphics[width=\linewidth]{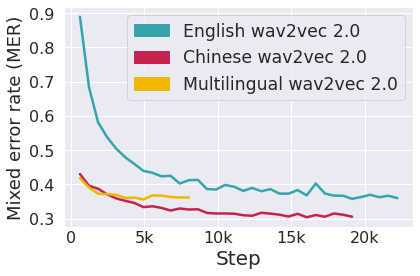}
    \captionof{figure}{MER on ASCEND validation set in the baseline experiments.}
    \label{fig:exp-mer}
\end{minipage}
\hfill
\begin{minipage}[b]{0.32\linewidth}
    \centering
    \includegraphics[width=\linewidth]{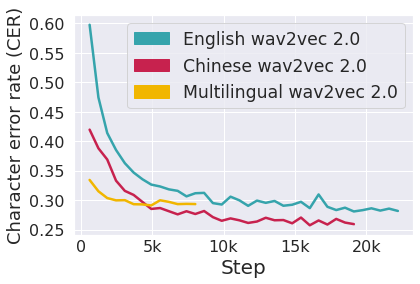}
    \captionof{figure}{CER on ASCEND validation set in the baseline experiments.}
    \label{fig:exp-cer}
\end{minipage}
\end{figure*}

For the audio data, we normalize the audio data and apply SpecAugment~\cite{Park2019} to increase the robustness of the model. Specifically, we apply time masking and frequency masking, with a time masking probability of 0.065, a time masking length of 2, a frequency masking probability of 0.004, and a frequency masking length of 2. No time warping is applied to the audio data.

\paragraph{Training details.} During the training, we employ Adam~\cite{Kingma2015AdamAM} to optimize the wav2vec 2.0 model. For the objective function, we use the Connectionist Temporal Classification (CTC) loss. The model is fine-tuned on a single GeForce GTX 3090 GPU with a learning rate of 5e-5 and a batch size of 16. We train the model up to 100 epochs with early stopping of 5 epochs.

\paragraph{Evaluation.} During the evaluation, we apply CTC decoding for generating the transcription. As for evaluation metrics, considering the character-based nature of the Chinese transcriptions and the word-based nature of the English transcriptions in ASCEND, we measure the models' performance using character error rate (CER) and mixed error rate (MER)~\cite{Fung2011SpeechRO,Hu2020,qiu20c_interspeech}. The CER is computed as the total of substitutions, deletions, and insertions divided by the number of characters in the reference, while the MER is calculated by measuring the CER for Chinese characters and word error rate (WER) for other characters.

% SpecAugment~\cite{Park2019} is a speech data augmentation method that works by warping the respective audio's features, masking a certain range of its time steps, and masking a certain range of its frequency channels.

\subsection{Result and analysis}

\begin{table}[t!]
    \centering
    \resizebox{1.0\linewidth}{!}{
        \begin{tabular}{cllll}
            \toprule
            \multirow{2}{2 cm}{\centering\textbf{Pre-training language}} & \multicolumn{2}{c}{\textbf{Validation}} & \multicolumn{2}{c}{\textbf{Test}} \\
                & \multicolumn{1}{c}{\textbf{MER (\%)}} & \multicolumn{1}{c}{\textbf{CER (\%)}} & \multicolumn{1}{c}{\textbf{MER (\%)}} & \multicolumn{1}{c}{\textbf{CER (\%)}} \\
            \midrule
            Chinese & \textbf{30.37} & \textbf{25.72} & \textbf{27.05} & \textbf{22.69} \\
            English & 35.77 & 28.07 & 28.72 & 22.78 \\
            Multilingual & 35.30 & 28.68 & 29.35 & 24.31 \\
            \bottomrule
        \end{tabular}
    }
    \caption{Baseline experiment results on ASCEND validation and test set. \textbf{Bold} denotes the best performance over different models.}
    \label{tab:exp-result}
\end{table}
The evaluation results of the English, Chinese, and multilingual pre-trained models are shown in Table \ref{tab:exp-result}. The experiment results suggest that the Chinese pre-trained model outperforms both the English and the multilingual pre-trained models. While the Chinese pre-trained model is slower at converging compared to the multilingual model, it converges much faster than the English one, as shown by the training loss curve in Figure \ref{fig:exp-train-loss}. Furthermore, Figure \ref{fig:exp-mer} and Figure \ref{fig:exp-cer} denote that the multilingual pre-trained model reaches a plateau earlier on both CER and MER in the ASCEND validation set. However, the Chinese pre-trained model ultimately yields better performance (30.37\% MER and 25.72\% CER) than the multilingual pre-trained model (35.30\% MER and 28.68\% CER) and the English pre-trained model (35.77\% MER and 28.07\% CER). This result is expected for three reasons: 1) almost 50\% of the language distribution in ASCEND is Chinese, 2) as presented by Table \ref{tab:vocab-size}, there is a huge vocabulary overlap between the Chinese pre-trained model and ASCEND-specific vocabulary, and 3) its pre-training solely focuses on learning Chinese instead of multiple languages at once like the multilingual model, in which the Chinese and English language only make up 0.95\% and 28.48\% of all the pre-training audio data.

Compared to other works on code-switching datasets~\cite{banerjee2018dataset,chowdhury2021omra,lynn2019irish,lyu2010seame,winata2020mtl}, the baseline experiment on ASCEND yields a comparable performance with $\sim$28\% MER and $\sim$23\% CER on the test set. Additionally, in terms of dataset size, the number of tokens, and word distribution, ASCEND is on par with other existing Chinese-English spontaneous code-switching datasets, such as CECOS~\cite{shen2011cecos} and SEAME~\cite{lyu2010seame}. These results indicate that ASCEND is reliable for training and evaluating Chinese-English code-switching ASR.

\section{Conclusion}

In this paper, we introduce ASCEND, a spontaneous multi-turn conversational dialogue Chinese-English code-switching corpus. ASCEND consists of 10.62 hours of spontaneous speech with a total of $\sim$12.3K utterances. The corpus is split into three sets: training, validation, and test with a ratio of 8:1:1 and a balanced gender proportion on each set. We further conduct a deeper analysis of the speech data to show the statistical distribution of both inter-sentential and intra-sentential code-switching utterances in ASCEND. Lastly, we experiment with the Chinese pre-trained wav2vec 2.0 model, English pre-trained wav2vec 2.0 model, and the multilingual pre-trained wav2vec 2.0 model to establish some baselines on ASCEND. Based on our experiment, the Chinese pre-trained model achieves the best code-switching performance (22.69\% CER and 27.05\% MER) on ASCEND's test set.

\section{Acknowledgements}

This work is funded by ITS/353/19FP of the Innovation Technology Commission, The Hong Kong SAR Government, School of Engineering Ph.D. Fellowship Award, The Hong Kong University of Science and Technology, and the Hong Kong Fellowship Scheme from the Hong Kong Research Grants Council (RGC).

\section{References}
\label{reference}
\label{sec:ref}

\bibliographystyle{lrec}
\bibliography{custom}

% \section{Language Resource References}
% \label{lr:ref}
% \bibliographystylelanguageresource{lrec}
% \bibliographylanguageresource{languageresource}

\end{document}